# New inference strategies for solving Markov Decision Processes using reversible jump MCMC

Matt Hoffman, Hendrik Kueck, Nando de Freitas, Arnaud Doucet
{hoffmanm,kueck,nando,arnaud}@cs.ubc.ca
University of British Columbia, Computer Science
Vancouver, BC, Canada

## Abstract

In this paper we build on previous work which uses inferences techniques, in particular Markov Chain Monte Carlo (MCMC) methods, to solve parameterized control problems. We propose a number of modifications in order to make this approach more practical in general, higher-dimensional spaces. We first introduce a new target distribution which is able to incorporate more reward information from sampled trajectories. We also show how to break strong correlations between the policy parameters and sampled trajectories in order to sample more freely. Finally, we show how to incorporate these techniques in a principled manner to obtain estimates of the optimal policy.

## 1 Introduction

The inference and learning approach for solving stochastic control and planning tasks, pioneered by many scientists (Dayan and Hinton, 1997; Attias, 2003; Doucet and Tadic, 2004; Verma and Rao, 2006; Toussaint and Storkey, 2006; Toussaint et al., 2006; Peters and Schaal, 2007), has enjoyed substantial success in the field of robotics (Toussaint et al., 2008; Kober and Peters, 2008; Hoffman et al., 2009; Vijayakumar et al., 2009). A significant body of empirical evidence in these papers also indicates that these methods can often outperform traditional stochastic planning and control methods, as well as more recent policy gradient schemes.

Notably, with the exception of (Doucet and Tadic, 2004; Hoffman et al., 2007a), the proposed algorithms have all been variants of the expectation-maximization (EM) procedure; see (Toussaint and Storkey, 2006) for a description of the EM approach. In the E step, a belief propagation algorithm is used to estimate the marginal distributions of the latent variables. To obtain analytical expressions for the belief updates, researchers following this line of work have been forced to truncate the time horizon of the Markov decision process (MDP). Moreover, they have had to focus on either discrete or linear-Gaussian models; where an abundant sea of alternatives already exists. Despite this, the EM approach was a key initial step in the development of more sophisticated inference and learning techniques for fully and partially observed MDPs.

These new inference and learning methods for stochastic control and planning have led to progress along several frontiers. First, the new methods apply naturally to structured MDPs, such as dynamic Bayesian network controllers, and hence can benefit from a rich body of knowledge in the fields of statistical inference and factored representations (Toussaint et al., 2008).

Second, the inference and learning approach opens up room for looking at the problem from a new angle, with the potential of leading to new insights and fostering innovation. For example, this strategy enabled (Hoffman et al., 2009) to relax the quadratic cost constraint in the classical and ubiquitous linear quadratic controllers via an EM algorithm with a tractable E step, but with general error functions. In that work, the authors also show that such an algorithm is equivalent to a policy gradient method where the simulation component is replaced by exact and efficient computations.

Third, the adoption of Monte Carlo EM has allowed for the extension of these methods to more general (hybrid discrete-continuous, nonlinear, non-Gaussian) state and action spaces (Hoffman et al., 2007a,b; Kober and Peters, 2008). The demonstrations with robots in (Kober and Peters, 2008) are particularly impressive. However, as noted in (Hoffman et al., 2007a), the forward-backward procedure in the Monte Carlo E step can be very computationally expensive. To avoid this, (Kober and Peters, 2008) use only forward simulation. This is an adequate pragmatic solution provided that the state space is not high-dimensional and/or that the rewards are not rare events (e.g. fairly flat reward functions with occasional narrow peaks). We note however that with increasing dimension, the probability of hitting a high reward (for many reward functions used in practice) becomes exponentially small.

To attack larger dimensional spaces with potentially rare rewards, (Hoffman et al., 2007a) proposes to sample from a trans-dimensional target distribution that is proportional to the reward function. Hence, the samples are guided to areas of high reward. This target distribution arises from a formulation of the stochastic control problem as one of carrying out Bayesian inference for an infinite mixture of finite-horizon MDPs, where the reward occurs at the last step of each MDP. A reversible jump MCMC algorithm (Green, 1995) is then developed to draw the states, actions, policy parameters and horizon lengths from the trans-dimensional target distribution. By doing this, they overcome the need for truncating the MDP that exists with the EM approach. They show that their proposed MCMC algorithm outperforms various Monte Carlo EM and policy search methods on a simple simulation example.

In this paper we propose several modifications and methodology improvements over (Hoffman et al., 2007a) that we believe are essential to make the method practical in more general, higher-dimensional state and policy spaces. First, we derive a new trans-dimensional target distribution that incorporates the rewards at all time steps. The need for the reward to occur only in the last time step of each mixture component was essential for deriving efficient two-filter smoothers in discrete MDPs (Toussaint and Storkey, 2006). The approach in (Hoffman et al., 2007a) inherited this restrictive requirement. However, we point out here that this requirement can be eliminated when adopting reversible jump MCMC algorithms. Within the MCMC framework, it is possible to introduce more informative target distributions that incorporate all the rewards without deteriorating the computational efficiency of the algorithm. In fact, as we show in the next subsection and the experimental section, the new target distributions can result in more efficient control algorithms.

Second, we note that the correlations among the variables of MDPs are very strong because of the natural time dependencies in these models. These correlations are one of the main causes of poor mixing times in high-dimensions. To overcome this fundamental problem, we propose a reformulation of the target distribution in terms of independent noise variables and deterministic transformations. This not only leads to huge improvements in mixing time, but also allows for the adoption of the common random numbers technique for variance reduction, which has been shown to perform well in control tasks (Ng and Jordan, 2000).

Our third major contribution addresses the problem of obtaining point estimates of the policy parameters. Whereas (Hoffman et al., 2007a) leaves the question of choosing a point estimate open, we present a strategy for finding these estimates by optimizing the marginal distribution of the policy parameters. This improvement has its roots in an elegant method developed for myopic Bayesian experimental design (Müller, 1998; Müller et al., 2004).

## 2 Description of the Model

We will assume a controlled Markov process over states, actions, and rewards $(X_n, U_n, R_n)$ evolving according to

- an initial state model: $x_0 \sim \mu(x_0)$,
- a parameterized policy: $u_n \sim p(u_n|x_n, \theta)$,
- a transition model: $x_{n+1} \sim p(x_{n+1}|x_n, u_n)$,
- and deterministic rewards: $r_n = r(x_n, u_n)$.[1]

In this paper we will be working in a model-based framework, in which these models are assumed to be known. Our goal is then to find the policy parameters $\theta^*$ which maximize the discounted expected reward

$$J(\theta) = \mathbb{E}[\sum_{n=0}^{\infty} \gamma^n \, r(Z_n)],$$

where $Z_n = (X_n, U_n)$ is the state/action pair at time $n$ and $\gamma < 1$ is a discount factor.

Following (Toussaint and Storkey, 2006), it is then possible to rewrite the expected reward as the expectation

$$J(\theta) = (1-\gamma)^{-1} \, \mathbb{E}[r(Z_K)],$$

which is taken with respect to the trans-dimensional distribution

$$p(k, z_{0:k}|\theta) = (1-\gamma) \, \gamma^k \, p(z_0|\theta) \prod_{n=1}^{k} p(z_n|z_{n-1}, \theta). \quad (1)$$

Here, the marginal distribution over trajectory lengths corresponds to the term $p(k) = (1-\gamma)\gamma^k$. Building on this result, (Hoffman et al., 2007a) introduced a reversible jump MCMC algorithm to produce samples of $(k, z_{0:k}, \theta)$ from a distribution with density

$$\bar{p}(k, z_{0:k}, \theta) \propto r(z_k) \, p(k, z_{0:k}|\theta) \, p(\theta), \quad (2)$$

where $p(\theta)$ is some prior distribution over $\theta$ chosen to ensure that our distribution is proper. We should note here that the use of this density requires a positive reward model, however this is easy to ensure so long as the rewards are bounded. If the integral $\int J(\theta)\, d\theta$ is finite we can choose an improper prior $p(\theta) \propto 1$. Another reasonable choice is a uniform distribution on a bounded region of the parameter space. This restricts the search for $\theta^*$ to a fixed region but does not alter the expected reward surface inside that region.

---

[1] More generally the rewards could be stochastic, but we will assume otherwise in order to simplify later notation.

By construction the target distribution (2) admits a marginal $\bar{p}(\theta) \propto J(\theta)\, p(\theta)$ as desired. Note that only the reward at the last time step of the MDP appears in this distribution. In the following subsection, we will show that it is possible to derive an alternative distribution that accounts for the sum of rewards at all time steps. The experiments will show that this new distribution has desirable properties.

### 2.1 Summing over all rewards

As mentioned above, the distribution utilized in (Hoffman et al., 2007a) and given by Equation 2 relies on weighting by rewards obtained at the end of the sampled trajectory. This formulation was crucial to the development of EM-based procedures such as (Toussaint and Storkey, 2006) in order to obtain efficient recursions. In the case of sampling-based procedures, however, we can greatly improve upon earlier methods by instead using the entire sequence of rewards $r_{0:k}$. We can first rearrange those terms which depend on the discount factor $\gamma$,

$$J(\theta) = (1-\gamma)^{-1} \sum_{n=0}^{\infty} \int r(z_n)\, p(n, z_{0:n}|\theta)\, dz_{0:n}$$

$$= (1-\gamma) \sum_{n=0}^{\infty} \frac{\gamma^n}{1-\gamma} \int r(z_n)\, p(z_{0:n}|n, \theta)\, dz_{0:n}$$

using the series expansion of $\gamma^n/(1-\gamma)$ we can write

$$= (1-\gamma) \sum_{n=0}^{\infty} \left( \sum_{k=n}^{\infty} \gamma^k \right) \int r(z_n)\, p(z_{0:n}|n, \theta)\, dz_{0:n}$$

finally, by changing the order of summation we obtain

$$= (1-\gamma) \sum_{k=0}^{\infty} \gamma^k \sum_{n=0}^{k} \int r(z_n)\, p(z_{0:n}|n, \theta)\, dz_{0:n}$$

$$= \sum_{k=0}^{\infty} \int \left( \sum_{n=0}^{k} r(z_n) \right) p(k, z_{0:k}|\theta)\, dz_{0:k}.$$

This reformulation allows us to introduce the new target distribution

$$\widetilde{p}(k, z_{0:k}, \theta) \propto \left( \sum_{n=0}^{k} r(z_n) \right) p(k, z_{0:k}|\theta)\, p(\theta), \quad (3)$$

which just as before results in a marginal proportional to the expected reward.

### 3 Explicit noise variables

While theoretically sound, sampling from (2) or (3) requires a carefully tuned proposal distribution in order to accurately explore the posterior. In many cases the policy parameters $\theta$ and the sequence of state/action pairs $z_{0:k}$ (as well as the individual steps within that sequence) will be highly correlated, resulting in a Markov chain which mixes very poorly over these variables. Blocking the variables can improve the mixing time of the Markov chain. Here, however, we adopt an even more efficient sampling strategy. In many models both the transition model and the policy take the form of deterministic functions with added noise[2], i.e.

$$u_n = \pi_\theta(x_n) + \phi_n \text{ and } x_{n+1} = f(x_n, u_n) + \psi_{n+1}.$$

where $\epsilon_n = (\phi_n, \psi_n)$ denotes the noise (i.e. stochastic) components which are distributed according to

$$p(\epsilon_n|\theta) = p(\psi_n)\, p(\phi_n|\theta).$$

Under this distribution the initial-state is a special case, however it becomes notationally convenient to consider this as "fully stochastic" and write $\psi_0 = x_0$. Here we allow the noise $\phi_n$ to depend upon $\theta$ so that the policy can control exploratory noise. In more general settings it might also make sense to let $x_0$ depend upon $\theta$, but here we assume that the initial state is uncontrolled.

Under these circumstances, the strong correlation exhibited by $z_n$ and $z_{n+1}$ is mostly due to the deterministic components. We remind the reader that it is this strong correlation that causes any MCMC algorithm to mix poorly. We can limit this problem by sampling $\epsilon_{0:k}$ rather than the state/path terms. This is an idea closely related to the techniques discussed by (Papaspiliopoulos et al., 2003). Under this re-parameterization, the new target distribution is

$$\widetilde{p}(k, \epsilon_{0:k}, \theta) \propto \left( \sum_{n=0}^{k} r(z_n) \right) p(k)\, p(\epsilon_{0:k}|k, \theta). \quad (4)$$

Although we have eliminated the need to sample $z_n$, we must still calculate it in order to compute the reward at time $n$; *this calculation is, however, deterministic given $z_{n-1}$ and $\epsilon_n$*.

The reformulation mitigates the mixing problems due to the strong dependencies between $\theta$ and $z_{0:k}$ as well as between $z_{n+1}$ and $z_n$ that were caused by the deterministic components of $p(u_n|x_n, \theta)$ and $p(x_{n+1}|x_n, u_n)$. The dependencies between the variables in this new artificial joint distribution are purely

---

[2] In this paper we present additive noise models purely for ease of exposition. It is trivial to generalize this approach to models where states are given by $x_{n+1} = f(x_n, u_n, \psi_{n+1})$ and actions by $u_n = \pi_\theta(x_n, \phi_n)$ for any functions $f$ and $\pi_\theta$.

due to the reward function $r$ and in many cases will be comparatively weak.

Apart from its decorrelating effect, this technique has a secondary benefit as a variance reduction technique. The noise terms $\epsilon_{0:k}$ can act as *common random numbers*, in a way that is closely related to the idea of fixing random seeds in policy search (i.e. PEGASUS (Ng and Jordan, 2000)). In particular, we can fix the noise variables for a predetermined number of MCMC moves updating the policy. In doing this, both $\theta_n$ and $\theta_{n-1}$ will depend on the same random seeds (noise terms). Consequently, the variance of the policy update will be reduced. This is a direct consequence of the fact that the variance of the difference of two estimators based on Monte Carlo simulations is equal to the sum of the individual variances of each estimator minus their joint covariance (Spall, 2005).

## 4  Marginal Optimization

### 4.1  Annealing

So far we have defined a distribution $\widetilde{p}(k, z_{0:k}, \theta)$ (Equation (3)) for which the policy parameter $\theta$ is marginally distributed as $J(\theta)\,p(\theta)$. Our goal however is to estimate the maximum $\theta^* = \arg\max_\theta J(\theta)$.

If $J(\theta)$ happens to have a strongly dominant and highly peaked mode around the global maximum $\theta^*$, we can justify sampling from $\widetilde{p}(k, z_{0:k}, \theta)$ and deriving a point estimate of $\theta^*$ by averaging the samples' $\theta$ components. This is the approach taken in (Hoffman et al., 2007a). However, in general the assumption of such a favorable $J(\theta)$ is unrealistic. If $J(\theta)$ is multimodal or fairly flat then this approach will yield poor estimates.

Instead, let us consider $\widetilde{p}_\nu(\theta) \propto J(\theta)^\nu p(\theta)$. For large exponents $\nu$ the probability mass of this distribution will concentrate on the global maximum $\theta^*$. If we could sample from $\widetilde{p}_\nu(\theta)$, then the generated samples would allow us to derive a much better point estimate of $\theta^*$. Note however that this is not as simple as it might seem at first glance. For example raising the joint density in Equation (3) to the power of $\nu$ will not result in a distribution with this desired marginal.

A method for generating samples from marginal distributions of this form was proposed in (Müller, 1998; Müller et al., 2004) in the context of optimal design and independently in (Doucet et al., 2002) in the context of marginal maximum a posteriori estimation. The trick is to define an artificial distribution jointly over multiple simulated trajectories. To simplify notation let us first define $\zeta_j = \{k_j, z_{0:k_j}\}$ to represent one simulated trajectory. Furthermore we will denote the accumulated (non discounted) reward along one simulated trajectory as $R(\zeta_j) = \sum_{n=0}^{k_j} r(z_n)$.

The appropriate artificial target distribution is then

$$\widetilde{p}_\nu(\theta, \zeta_{1:\nu}) \propto p(\theta) \prod_{j=1}^{\nu} R(\zeta_j)\, p(\zeta_j|\theta), \qquad (5)$$

where $\nu$ is a positive integer and $p(\zeta_j|\theta)$ is given by Equation (1). It is easy to verify that this distribution does indeed admit the desired marginal distribution $\widetilde{p}_\nu(\theta) \propto J(\theta)^\nu p(\theta)$. However, because the modes of $J(\theta)^\nu$ will typically be narrow and widely separated for large $\nu$, sampling from this distribution using Markov chain Monte Carlo techniques directly is difficult.

Therefore, we take a simulated annealing approach (as in (Müller, 1998; Müller et al., 2004; Doucet et al., 2002)) in which we start sampling from $\widetilde{p}_\nu$ with $\nu=1$, and then slowly increase $\nu$ over time according to an annealing schedule. Increasing $\nu$ slowly enough allows the chain to efficiently explore the whole parameter space before becoming more constrained to the major modes for larger values of $\nu$.

One limitation of Equation (5) is that the annealing schedule is limited to full integer steps. However, we can further generalize this approach to allow for a real valued annealing schedule by defining the modified target distribution

$$\widetilde{p}_\nu(\theta, \zeta_{1:\lceil\nu\rceil}) \propto \\ p(\theta) \left( \prod_{j=1}^{\lfloor\nu\rfloor} R(\zeta_j)\, p(\zeta_j|\theta) \right) R(\zeta_{\lceil\nu\rceil})^{\nu-\lfloor\nu\rfloor}\, p(\zeta_{\lceil\nu\rceil}|\theta), \quad (6)$$

where $\nu$ is now real valued and $\lfloor\nu\rfloor$ and $\lceil\nu\rceil$ denote the integer valued floor and ceiling of $\nu$.

For integer values of $\nu$, this distribution again admits the marginal $\widetilde{p}_\nu(\theta) \propto J(\theta)^\nu p(\theta)$ as before. While this does not hold for the intermediate distributions with real valued $\nu$, these distributions provide a smooth bridge between the integer steps. This allows for more gradual annealing, thereby reducing the variance of the overall sampler.

While in theory we should let $\nu$ approach infinity, in practice this is not computationally feasible. Instead we choose an annealing schedule that plateaus at a final integer value $\nu_{\max}$, at which point the chain is run for another $M$ iterations. These last $M$ samples from $\widetilde{p}_{\nu_{\max}}(\theta)$ are then used as the basis of a point estimate of $\theta^*$.

### 4.2  Clustering

If $J(\theta)$ has a unique maximum and $\nu_{\max}$ is sufficiently large, the final samples from $p_{\nu_{max}}(\theta)$ will all be concentrated around $\theta^*$. In this case averaging the $L$ final

**Require:** an initial sample $\theta^{(0)}$
1: initialize our trajectories with $\zeta_0$ given $\theta^{(0)}$
2: initialize $\nu := 1$
3: **for** $i = 1$ to $N$ **do**
4: 　**for all** trajectories $\zeta_j$ **do**
5: 　　perform a *birth or death* move on $\zeta_j$
6: 　　if $i$ is divisible by $n_{\text{up}}$
　　　　perform an *update* move on $\zeta_j$
7: 　**end for**
8: 　sample $\theta^{(i)}$ given $\theta^{(i-1)}$
9: 　update $\nu$ with some annealing schedule
10: 　sample a new trajectory $\zeta^{(i)}_{\lceil\nu\rceil}$ if necessary
11: **end for**
12: cluster all $\{\theta^{(i)}\}$ obtained under $\nu_{\max}$

Figure 1: The reversible-jump MDP algorithm.

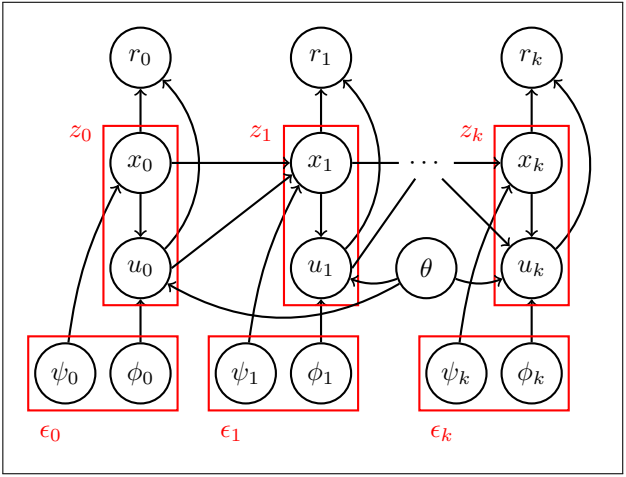

Figure 2: A graphical model depicting the interactions between state/action, noise, and reward terms (as well as any dependency on $\theta$).

samples can provide a good estimate of $\theta^*$. In practice however, it is possible that $p_{\nu_{max}}$ still has multiple modes with significant probability mass. In this case simple averaging can lead to a poor estimate.

To provide a better point estimate under these circumstances we cluster the final samples and use the center of the largest cluster as our estimate of $\theta^*$. For the clustering we use simple agglomerative clustering using average linkage (UPGMA). Other techniques such as for example mean shift clustering (Cheng, 1995) could be used instead. Note however that the popular K-Means algorithm is not suited for this purpose as it tends to split high density areas into multiple clusters.

## 5　Algorithm

While we cannot directly sample from the annealed distribution in (6), a Markov chain targeting this distribution can be constructed. For a given $\theta$ a combination of birth, death, and blocked update moves are used to propose updates for each trajectory in $\{\zeta_j\}$. The birth and death moves allow us to mix across trajectories of different dimensions, and it is here that we need to bring in the machinery of reversible jump MCMC. Finally, given the trajectory samples we can update $\theta$ using Metropolis-Hastings. Acceptance probabilities for these various moves are described later, and pseudo-code for the algorithm is given in Figure 1.

Before describing the acceptance ratios for this algorithm, however, we should first discuss the relationship between the different trajectory terms. Although the reward and state/action terms are deterministic given the noise variables, we can see from Equation (6) that they must still be computed in order to evaluate the target density (up to a normalizing constant). As noted before, though, modifying $\epsilon_n$ or inserting a new term $\epsilon_n^*$ requires that we recalculate all following state, action, and reward terms—a fact that can be readily verified via the graphical model shown in Figure 2. In this work the fact that we are working directly with the noise variables means that we are updating the actual states $z_n$ only indirectly. Birth/death moves at intermediate points in a trajectory would shift the sequence of random variates, which effectively combines the birth/death move with a potentially high dimensional blocked update move. This would lead to lower acceptance rates and worse mixing of the chain in addition to higher computational cost. We therefore only sample birth and death moves at the end of a trajectory (i.e. for $n = k$) and update intermediate noise terms using the largest block size for which the acceptance ratios remain reasonable. Using birth/death moves only at the end of the trajectory is theoretically valid because the update moves eventually update all states.

At every iteration and for each trajectory we will propose a new trajectory $\zeta_j^*$ given $\zeta_j$ and $\theta$. The trajectories are conditionally independent given $\theta$, and as a result we can write the acceptance probability for each proposed move as $\min(1, \alpha)$, where

$$\alpha = \frac{\widetilde{p}_\nu(\zeta_j^*|\theta)}{\widetilde{p}_\nu(\zeta_j|\theta)} \cdot \frac{q(\zeta_j|\zeta_j^*)}{q(\zeta_j^*|\zeta_j)}$$

for some proposal density $q$. Since each trajectory can be sampled independently we will drop the $j$ indices and write the internal variables for each trajectory as $\zeta = (k, \epsilon_{0:k}, z_{0:k})$. We will also define the "annealing exponent" for each trajectory, as $\beta = \nu - \lfloor \nu \rfloor$ if $\zeta$ is the final trajectory and $\beta = 1$ otherwise.

At each iteration we propose a birth move with probability $b_k$ or a death move with probability $d_k = 1 - b_k$.[3] Proposing a birth move involves sampling a new noise term $\epsilon_{k+1}^*$ and calculating $z_{k+1}^*$; we accept with probability $\min(1, \alpha_{\text{birth}})$, where

$$\alpha_{\text{birth}} = \frac{p(k+1)}{p(k)} \frac{d_{k+1}}{b_k} \frac{p(\epsilon_{1:k})\, p(\epsilon_{k+1}^*)}{p(\epsilon_{1:k})\, q(\epsilon_{k+1}^*)} \left[\frac{R(z_{0:k}, z_{k+1}^*)}{R(z_{0:k})}\right]^\beta$$
$$= \gamma \frac{d_{k+1}}{b_k} \left[\frac{R(z_{0:k}, z_{k+1}^*)}{R(z_{0:k})}\right]^\beta.$$

The simplification in this last step is due to the fact that we are able to sample directly from the noise model, and as a result $q = p$. If a death move is proposed we need only remove the last noise and state/action term and hence there is no need to sample. The acceptance ratio for this move can be obtained by inverting the birth move ratio, and as a result we will accept with probability $\min(1, \alpha_{\text{death}})$, where

$$\alpha_{\text{death}} = \frac{1}{\gamma} \frac{b_{k-1}}{d_k} \left[\frac{R(z_{0:k-1})}{R(z_{0:k})}\right]^\beta.$$

Every $n_{\text{up}}$ iterations a fixed-dimensional update move is proposed. We will first randomly select a block of $n_{\text{block}}$ variables $[a, b]$. Given this block we can sample new noise terms $\epsilon_{a:b}^*$ and calculate the corresponding path terms $z_{a:b}^*$ and $z_{b+1:k}^*$. The move will be accepted with probability $\min(1, \alpha_{\text{update}})$, where

$$\alpha_{\text{update}} = \frac{p(\epsilon_{a:b}^*)\, q(\epsilon_{a:b})}{p(\epsilon_{a:b})\, q(\epsilon_{a:b}^*)} \left[\frac{R(z_{0:a-1}, z_{a:b}^*, z_{b+1:k}^*)}{R(z_{0:a-1}, z_{a:b}, z_{b+1:k}^*)}\right]^\beta$$
$$= \left[\frac{R(z_{0:a-1}, z_{a:b}^*, z_{b+1:k}^*)}{R(z_{0:k})}\right]^\beta.$$

Finally, given the trajectories $\{\zeta_j\}$ we can sample a new set of policy parameters $\theta^*$ from the proposal distribution $q(\theta^*|\theta)$. This new policy requires us to recalculate the state/action components for each trajectory, resulting in $\zeta_j^*$. These new policy parameters are then accepted with probability $\min(1, \alpha_{\text{mh}})$ where

$$\alpha_{\text{mh}} = \frac{\prod_{j=0}^{\lfloor \nu \rfloor} R(\zeta_j^*)\, p(\zeta_j^*|\theta^*)}{\prod_{j=0}^{\lfloor \nu \rfloor} R(\zeta_j)\, p(\zeta_j|\theta)}$$
$$\frac{R(\zeta_{\lceil \nu \rceil}^*)^{\nu - \lfloor \nu \rfloor}\, p(\zeta_{\lceil \nu \rceil}^*|\theta^*)}{R(\zeta_{\lceil \nu \rceil})^{\nu - \lfloor \nu \rfloor}\, p(\zeta_{\lceil \nu \rceil}|\theta)} \cdot \frac{q(\theta|\theta^*)}{q(\theta^*|\theta)} \frac{p(\theta^*)}{p(\theta)}.$$

This form of the acceptance ratio is completely general, however a number of assumptions can be made in practice which greatly simplify its form. A common choice for the proposal $q$ is a symmetric distribution such that $q(\theta, \theta') = q(\theta', \theta)$. If the proposal distribution is symmetric and the prior $p(\theta)$ is uniformly distributed, as noted in Section 2, then the final two terms of this acceptance ratio will cancel. Finally, if additionally the policy noise $\phi_{0:k}$ is independent of $\theta$, or if the policy is deterministic, the acceptance ratio simplifies to

$$\alpha_{\text{mh}} = \frac{\left[\prod_{j=0}^{\lfloor \nu \rfloor} R(\zeta_j^*)\right] R(\zeta_{\lceil \nu \rceil}^*)^{\nu - \lfloor \nu \rfloor}}{\left[\prod_{j=0}^{\lfloor \nu \rfloor} R(\zeta_j)\right] R(\zeta_{\lceil \nu \rceil})^{\nu - \lfloor \nu \rfloor}}.$$

---

[3]In general we will let these terms be constant for all $k$, but it must be the case that $b_0 = 1$ in order to ensure that we do not "kill off" chains of length one.

If there is exploitable structure in $\theta$ we can also propose blocked updates of the parameters.

## 6 Experiments

### 6.1 Linear-Gaussian models

We first experiment with linear-Gaussian transition models of the form

$$f(x_n, u_n) = Ax_n + Bu_n + \mathcal{N}(0, \Sigma), \text{ and}$$
$$\pi_\theta(x_n) = Kx_n + m \text{ for } \theta = (K, m).$$

This model is particularly interesting if we allow for multimodal rewards, as this will in general induce a multimodal expected reward surface. Figure 3 contrasts samples taken from both the non-annealed and annealed distribution (with annealing factor $\nu = 20$) on a model with 1D state- and action-spaces. In this example we can see that the simple approach of averaging samples $\{\theta^{(i)}\}$ results in a very poor estimate of the policy parameters, whereas both clustering and annealing are correctly able to recover the optimum.

### 6.2 Particles with force-fields

For a more challenging control problem we chose to simulate a physical system in which a number of repellers are affecting the fall of particles released from within a start region. The goal is to direct the path of the particles through high reward regions of the state space in order to maximize the accumulated discounted reward. The four-dimensional state-space in this problem consists of a particle's position and velocity $(p, \dot{p})$ for $p \in \mathbb{R}^2$. Actions consist of repelling forces acting on the particle. Additionally, the particle is affected by gravity and a frictional force resisting movement.

The deterministic policy is parameterized by $k$ repeller positions $r_i$ and strengths $w_i$ with a functional form

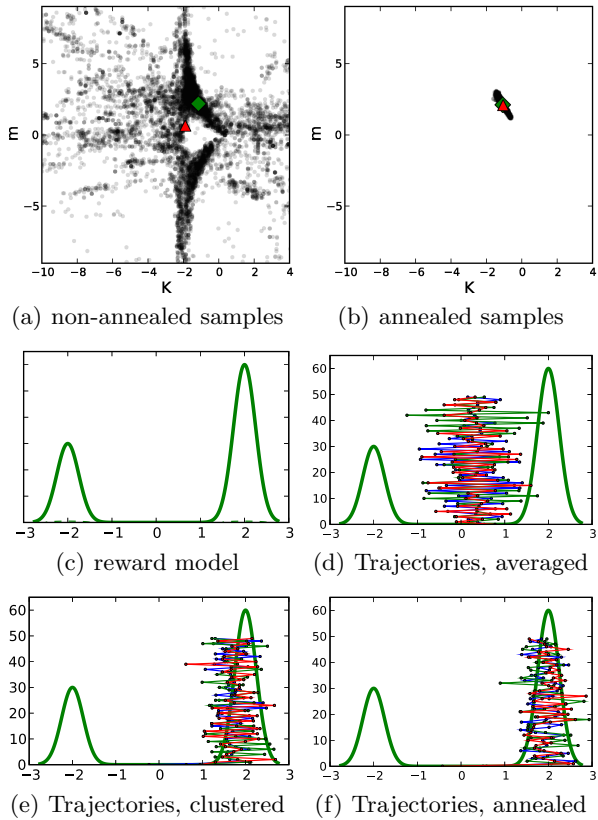

(a) non-annealed samples   (b) annealed samples

(c) reward model   (d) Trajectories, averaged

(e) Trajectories, clustered   (f) Trajectories, annealed

Figure 3: Linear-Gaussian example with multimodal reward, shown in (c). The top two plots show samples of the 2D policy parameters, where (a) displays those samples taken without annealing, and (b) those with annealing. Simple averaging of the sampled parameters in (a) leads to an estimate given by the red triangle, whereas the green diamond is the point estimate found by clustering these same samples. Also shown (d-f) are sample trajectories under these 3 different estimates where the y-axis gives the discrete time index.

given by

$$\pi_\theta(p, \dot{p}) = \sum_{i=1}^{k} w_i \frac{p - r_i}{\|p - r_i\|^3}.$$

That is, each repeller pushes the particles directly away from it with a force inversely proportional to its distance from the particle. In our experiments the particle's start position is uniformly distributed within a rectangular region (shown in yellow in Figures 4 and 5). At each time step the particle's position and velocity are updated using simple deterministic physical forward simulation and a small amount of Gaussian transition noise is added to the particle's velocity.

In Figure 4 we use this particle model to show the benefits of the proposed summed reward formulation (Equation (3)) over the target distribution which only uses rewards at the last time step (Equation (2)). We employ the noise variable parameterization and the annealing and clustering techniques discussed in Sections 3 and 4 in both samplers.

The reward model used in this example is composed of multiple circular reward zones. A high constant reward is awarded inside these zones and close to no reward outside. Note that the discontinuous and multimodal nature of this reward surface makes for a very challenging control problem. In this and the following experiments we are searching for the optimal placement and strengths of two repellers, resulting in a 6 dimensional control problem. In our implementation we are updating the 6 policy components in 4 blocks for the positions and strengths of the two repellers.

When evaluating the reward at the last step only, the sampler has difficulties crossing the gaps between the reward zones, as indicated by the relatively low acceptance ratios of birth and death moves, see Figure 4(b). This leads to the sampler getting stuck in local minima, resulting in poor policy estimates. The summed rewards formulation on the other hand allows for better mixing over path lengths, making it more likely to find the high reward zone at the bottom. This ultimately results in much better policies.

Note how the policy found using our summed rewards approach and visualized in Figure 4(d) uses the two repellers to not only direct the particles towards the high reward zone but to also slow them down inside this zone in order to accumulate as much reward as possible.

Figure 5 compares the algorithm described in this paper with the PEGASUS technique (Ng and Jordan, 2000) using numerically computed gradients. In particular we are interested in learning using deterministic policies, and PEGASUS can be used directly in this setting. We compared 10 runs of each algorithm on the particle system model shown in the bottom two subplots, where the reward model is a single Gaussian in position-space. Even though the reward model is unimodal, the resulting expected reward surface is highly multimodal: two such modes are displayed in the bottom two subplots. The poor performance and high variance of PEGASUS is mainly due to these local maxima, as well as plateaus in the reward surface.

Figure 6 uses the same problem from the previous experiment to compare the sampler based on the noise-variable formulation with the reversible jump approach used in (Hoffman et al., 2007a). By examining the resulting policy estimates we can see that the proposed reformulation significantly outperforms the previous method on this model. This results from the older method's poor mixing over trajectories, as evidenced by the extremely low acceptance rate for path updates. In order to explore the space of trajectories at all, this method therefore needs to shrink trajectories using death moves and subsequently re-grow them using birth moves. However, the acceptance ratios for such birth and death moves are themselves significantly lower than for our proposed sampler, rendering this way of mixing in trajectory space inefficient as well. The earlier work of (Hoffman et al., 2007a) was able to avoid these inefficiencies by using a hand crafted proposal mechanism that helped to break the state-space dependencies. This proposal was chosen in an ad hoc, model-dependent manner, however, and unlike our approach is not applicable to more general problems.

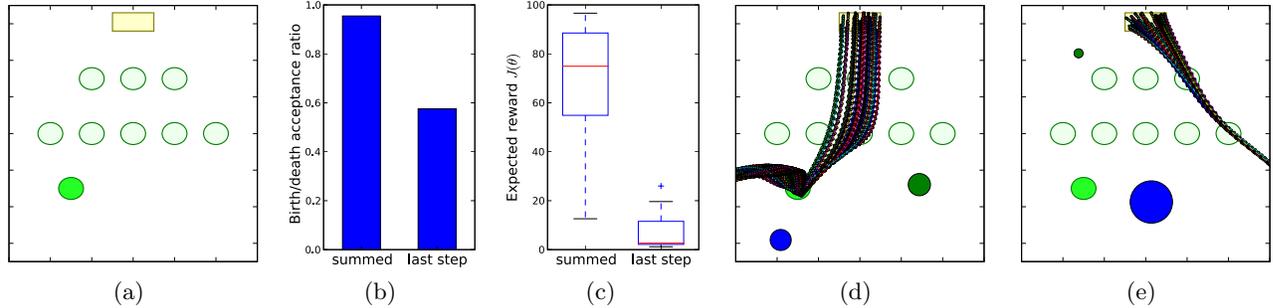

(a) (b) (c) (d) (e)

Figure 4: Comparison of a sampler evaluating the reward only at the last step of a simulated trajectory and our proposed sampler, which sums all rewards along the trajectory. The problem, shown in (a), features multiple reward zones, with the bottom-most zone yielding a 50 times higher reward than the others. The average acceptance ratios for both samplers are displayed in (b), while (c) compares the expected rewards for the policies found using 10 runs of each sampler. The final two plots visualize two of the computed policies; one for the summed reward formulation in (d) and one when only evaluating the reward at the last step in (e).

## 7 Conclusion

In this paper we have presented several important improvements to the approach of (Hoffman et al., 2007a) for solving general MDPs using reversible jump MCMC. The experiments provide clear evidence that the proposed modifications are needed to attack higher-dimensional stochastic decision problems. In particular, the experimental results show that significant improvements are obtained when incorporating more reward information (Figure 4) and when using the explicit noise variables to break state-space dependencies and reduce variance (Figure 6). It is also clear that the proposed simulated annealing and clustering techniques allow us to find better point estimates of the optimal policy (Figure 3). Finally, we observed favorable performance of the proposed approach in comparison to state-of-the-art techniques such as PEGASUS (Figure 5).

The repellers example used a 4-dimensional state space and 6-dimensional policy space. Scaling to higher dimensional state spaces should be possible in principle.

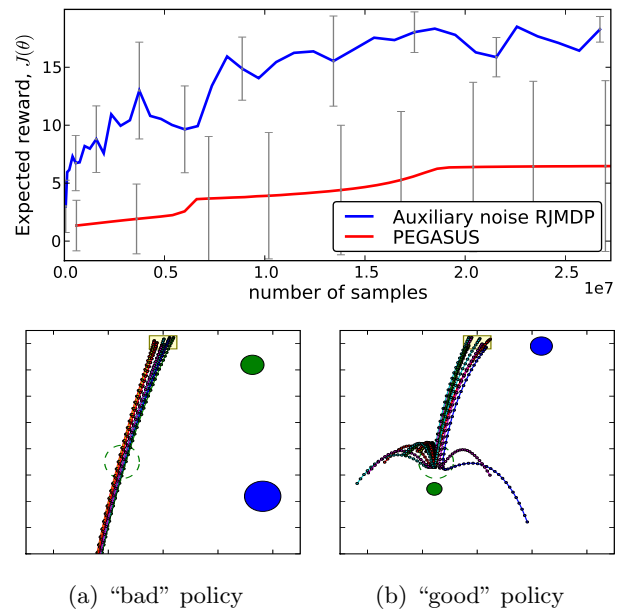

(a) "bad" policy (b) "good" policy

Figure 5: Comparison with PEGASUS on the repellers model, averaged over 10 runs, where error-bars display one standard deviation. The x-axis displays the number of samples taken from the transition model. Also shown are (a) a "bad" local maxima found by PEGASUS, and (b) a "good" policy found by our sampler.

As long as there is structure in the state space, one can adopt standard Rao-Blackwellization and blocking techniques to efficiently carry out inference in the Bayesian network. The main difficulty here lies in dealing with the dimensionality of the policy space, where often there seems to be much less structure to exploit. How to recruit more structure or gradients (when the model is differentiable) is an ongoing research direction.

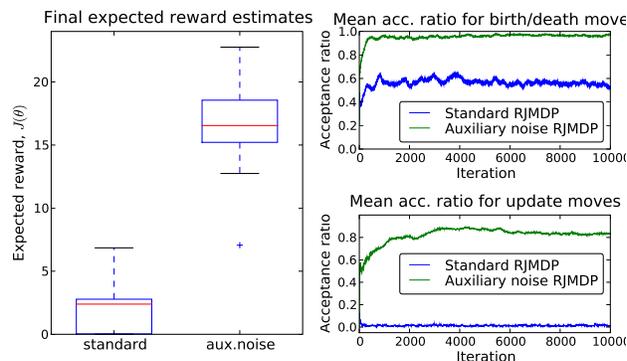

Figure 6: Comparison between the explicit noise variable approach and the standard approach of (Hoffman et al., 2007a) on the particle system model. The leftmost plot shows the expected rewards for the final policies found by both methods across 10 runs. The right plots display the averaged acceptance ratios for birth and death moves and the acceptance ratios for trajectory update moves.

## Acknowledgements

This work was supported by the Natural Sciences and Engineering Research Council of Canada (NSERC) and the Mathematics of Information Technology and Complex Systems (MITACS). M. Hoffman was also supported by a UBC Graduate Fellowship.

## References

H. Attias. Planning by probabilistic inference. In *UAI*, 2003.

Y. Cheng. Mean shift, mode seeking, and clustering. *IEEE Transactions on Pattern Analysis and Machine Intelligence*, 17(8):790–799, 1995.

P. Dayan and G. Hinton. Using EM for reinforcement learning. *Neural Computation*, 9:271–278, 1997.

A. Doucet and V. Tadic. On solving integral equations using Markov Chain Monte Carlo methods. Technical Report CUED-F-INFENG 444, Cambridge University Engineering Department, 2004.

A. Doucet, S. Godsill, and C. Robert. Marginal maximum a posteriori estimation using Markov chain Monte Carlo. *Statistics and Computing*, 12(1):77–84, 2002.

P. Green. Reversible jump Markov Chain Monte Carlo computation and Bayesian model determination. *Biometrika*, 82(4):711–732, 1995.

M. Hoffman, A. Doucet, N. de Freitas, and A. Jasra. Bayesian policy learning with trans-dimensional MCMC. In *NIPS*, 2007a.

M. Hoffman, A. Doucet, N. de Freitas, and A. Jasra. On solving general state-space sequential decision problems using inference algorithms. Technical Report TR-2007-04, University of British Columbia, Computer Science, 2007b.

M. Hoffman, N. de Freitas, A. Doucet, and J. Peters. An expectation maximization algorithm for continuous Markov Decision Processes with arbitrary reward. In *AI-STATS*, 2009.

J. Kober and J. Peters. Policy search for motor primitives in robotics. In *NIPS*, 2008.

P. Müller. Simulation based optimal design. In *Bayesian Statistics 6*, 1998.

P. Müller, B. Sansó, and M. de Iorio. Optimal Bayesian design by inhomogeneous Markov chain simulation. *Journal of the American Statistical Association*, 99: 788–798, 2004.

A. Ng and M. Jordan. PEGASUS: A policy search method for large MDPs and POMDPs. In *UAI*, pages 406–415, 2000.

O. Papaspiliopoulos, G. Roberts, and M. Sköld. Non-centered parameterisations for hierarchical models and data augmentation. *Bayesian Statistics*, 7, 2003.

J. Peters and S. Schaal. Reinforcement learning for operational space control. In *ICRA*, 2007.

J. Spall. *Introduction to stochastic search and optimization: estimation, simulation, and control*. Wiley-Interscience, 2005.

M. Toussaint and A. Storkey. Probabilistic inference for solving discrete and continuous state Markov Decision Processes. In *ICML*, 2006.

M. Toussaint, S. Harmeling, and A. Storkey. Probabilistic inference for solving (PO)MDPs. Technical Report EDI-INF-RR-0934, University of Edinburgh, School of Informatics, 2006.

M. Toussaint, L. Charlin, and P. Poupart. Hierarchical POMDP controller optimization by likelihood maximization. In *UAI*, pages 562–570, 2008.

D. Verma and R. Rao. Planning and acting in uncertain environments using probabilistic inference. In *IROS*, 2006.

S. Vijayakumar, M. Toussaint, G. Petkos, and M. Howard. Planning and moving in dynamic environments: A statistical machine learning approach. In Sendhoff, Koerner, Sporns, Ritter, and Doya, editors, *Creating Brain Like Intelligence: From Principles to Complex Intelligent Systems, LNAI-Vol. 5436*. Springer-Verlag, 2009.